\documentclass[letterpaper,twocolumn,fleqn]{article} 

\usepackage{ist}
\usepackage{amsmath}
\usepackage{xcolor}
\usepackage{graphicx}
\usepackage[hidelinks]{hyperref}
\usepackage{hyperref}
\usepackage{fontawesome}

\hypersetup{
    colorlinks=true,
    linkcolor=blue,   % Color of internal links
    citecolor=blue,   % Color of citation links
    urlcolor=red      % Color of external links
}

\ifx00 %ifx01 to turn off and ifx00 to turn on 
\newcommand{\SKS}[1]{\textcolor{red}{[Sushil: #1]}}
\else 
\newcommand{\SKS}[1]{\textcolor{red}{}}
\fi

\ifx00 %ifx01 to turn off and ifx00 to turn on 
\newcommand{\SC}[1]{\textcolor{blue}{[Sanjay: #1]}}
\else 
\newcommand{\SC}[1]{\textcolor{blue}{}}
\fi
\title{Minimizing Occlusion Effect on Multi-View Camera Perception in BEV with Multi-Sensor Fusion}

\author{ 
Sanjay Kumar\textsuperscript{1,2,3}, 
Hiep Truong\textsuperscript{1,4}, 
Sushil Sharma\textsuperscript{1,2}, 
Ganesh Sistu\textsuperscript{1,5}, 
Tony Scanlan\textsuperscript{1,2}, 
Eoin Grua\textsuperscript{1,2}, 
Ciarán Eising\textsuperscript{1,2,3}\\
\textsuperscript{1} Department of Electronic and Computer Engineering, University of Limerick, Ireland\\
\textsuperscript{2} Data-Driven Computer Engineering (D$^2$iCE) Research Centre, University of Limerick, Ireland\\
\textsuperscript{3} Lero, The Irish Software Research Centre, University of Limerick, Ireland\\
\textsuperscript{4} DSW, Valeo Kronach, Germany\\
\textsuperscript{5} Valeo Vision Systems, Ireland}

\date{} % date has an empty field.

\begin{document} 

\maketitle 

\thispagestyle{empty} % prevents the first page to be numbered

%%%%%%%%%%%%%%%%%%%%%%%%%%%%%%%%%%
% Abstract
%%%%%%%%%%%%%%%%%%%%%%%%%%%%%%%%%%

\begin{abstract}
Autonomous driving technology is rapidly evolving, offering the potential for safer and more efficient transportation. However, the performance of these systems can be significantly compromised by the occlusion on sensors due to environmental factors like dirt, dust, rain, and fog. These occlusions severely affect vision-based tasks such as object detection, vehicle segmentation, and lane recognition. In this paper, we investigate the impact of various kinds of occlusions on camera sensor by projecting their effects from multi-view camera images of the nuScenes dataset into the Bird's-Eye View (BEV) domain. This approach allows us to analyze how occlusions spatially distribute and influence vehicle segmentation accuracy within the BEV domain. Despite significant advances in sensor technology and multi-sensor fusion, a gap remains in the existing literature regarding the specific effects of camera occlusions on BEV-based perception systems. To address this gap, we use a multi-sensor fusion technique that integrates LiDAR and radar sensor data to mitigate the performance degradation caused by occluded cameras. Our findings demonstrate that this approach significantly enhances the accuracy and robustness of vehicle segmentation tasks, leading to more reliable autonomous driving systems. \textsf{\href{https://youtu.be/OmX2NEeOzAE}{\url{https://youtu.be/OmX2NEeOzAE}}}\\

\end{abstract}

%%%%%%%%%%%%%%%%%%%%%%%%%%%%%%%%%%%%
% Overall Document Guidelines: Head
%%%%%%%%%%%%%%%%%%%%%%%%%%%%%%%%%%%%
\section{INTRODUCTION}
\begin{figure}[!t]
  \centering
  \includegraphics[width=0.99\columnwidth]{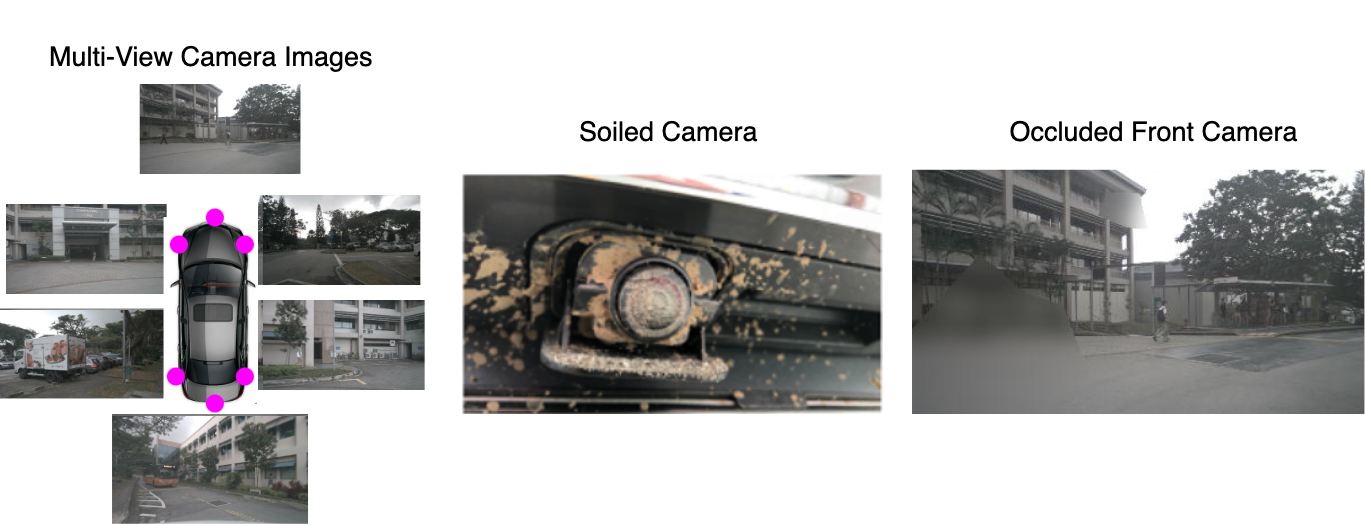}
  \caption{Left: Multi-view cameras and their mounting positions on the ego vehicle. Middle: Camera lens obstructed by mud, raindrops, dust, etc. Right: Occluded image}
  \label{Figure:Intro}
\end{figure}

Autonomous vehicles (AVs) and Advanced Driver Assistance Systems (ADAS) rely on a suite of sensors such as cameras, radar, and LiDAR to perceive and interpret their surroundings \cite{bib1,bib2}. These systems ensure the vehicles' safe and efficient navigation in complex driving environments \cite{bib3}. Among these sensors, cameras play a significant role due to their ability to capture high-resolution images enriched with semantic information. These images are essential for various perception tasks, including object detection \cite{bib4}, vehicle segmentation \cite{bib5}, and lane recognition. However, despite their importance, camera-based perception systems face significant challenges from environmental factors that can degrade visual data quality \cite{bib6}. One of the most critical issues in this context is camera occlusion, where contaminants like dirt, raindrops, or snow obstruct the camera lens, leading to a significant reduction in image clarity, demonstrated in Figure \ref{Figure:Intro}. This degradation can adversely affect the performance of vision-based algorithms, ultimately compromising the safety and reliability of AVs \cite{bib7}. As visual data is crucial for generating the semantic representation of the environment, any occlusion can have a cascading effect on the vehicle's perception capabilities. \\

Previous studies have addressed various challenges related to soiling (occlusions), sun glare, and the integration of multi-sensor data for autonomous driving \cite{bib8, bib9, bib10, bib11, bib12, bib13, bib14}. However, these studies have primarily focused on specific sensor types or isolated components of the perception system, leaving the impact of environmental factors on Bird’s-Eye View (BEV) transformations underexplored. While recent research has begun to examine multi-sensor BEV perception, the challenges posed by camera occlusion within this framework remain inadequately addressed.
In response, our paper uses a multi-sensor fusion technique incorporating LiDAR and radar data to counteract the effects of camera occlusion within \textbf{BEV-based perception systems}. Unlike previous research that primarily focused on alleviating occlusion effects on cameras, our approach emphasizes understanding and addressing the degradation within the BEV space. Our research demonstrates significant improvements in vehicle segmentation, addressing a gap in the current understanding of how environmental challenges impact BEV representations and overall AV performance. 

The primary contributions of this paper are as follows:

\begin{itemize}
    \item We create \textbf{artificial occlusion} on nuScenes \cite{bib16} multi-view cameras using woodscape soiling dataset Patterns \cite{bib17}. 
    \item We evaluate how camera occlusion affects \textbf{Bird’s Eye View (BEV)} perception tasks, particularly focusing on the resulting degradation in vehicle segmentation performance.
    \item We use \textbf{multi-sensor fusion technique} using simple-BEV \cite{bib15} that integrates radar and LiDAR with occluded camera inputs, improving vehicle segmentation accuracy.
    \item  Our study addresses a key research gap on camera occlusion's effects on BEV space.
\end{itemize}

This paper is organized as follows: We begin with a review of the research background, followed by the methodology, which details how the occlusion is applied to the nuScenes dataset and the architecture details. Next, we present the experimental results. Finally, we discuss important findings and suggestions for future work.

\section{RESEARCH BACKGROUND}

Numerous studies highlight the significant role of multi-sensor perception systems in advancing autonomous vehicles (AVs) and Advanced Driver Assistance Systems (ADAS). The Bird’s-Eye View (BEV) transformation \cite{bib18, bib19}, which integrates data from multiple sensors for tasks such as vehicle segmentation, object detection, and path planning, is particularly important \cite{bib18}. However, most of the existing research has primarily focused on detecting camera lens soiling, without thoroughly measuring the impact of environmental factors like rain, dust, and fog on perception algorithms. This gap underscores the need for further investigation into how these occlusions affect the performance of BEV systems.

\subsection{Research Gap}
Our research aims to investigate how occlusions such as dirt, fog, and raindrops affect BEV (Bird's Eye View) perception algorithms, with a specific focus on vehicle segmentation. By integrating data from radar and LiDAR sensors, we seek to overcome the degradation caused by these occlusions in the BEV domain. This integration will enhance the reliability of perception systems under occluded conditions, leading to improved performance in key tasks such as vehicle segmentation for autonomous vehicles.

\subsection{Soiling Impact on Camera Perception}
The effect of soiling on camera lenses was addressed using CycleGAN-based image restoration, where a De-soiling dataset was used to train the Generative Adversarial Network (GAN), resulting in improved road and lane detection in fisheye camera images \cite{bib9}. SoilingNet, a Convolutional Neural Network (CNN) designed to detect various soiling types on automotive cameras, utilized a multi-branch approach and GAN-based data augmentation to enhance robustness \cite{bib10}. While these approaches improved camera perception, they did not investigate the impact on BEV systems. Additionally, these studies did not explore the broader effects of soiling on perception algorithms, focusing solely on camera-specific improvements.

In a related study, sun glare was addressed using a Glare Detection Network (GDN), which demonstrated its effect on camera perception, although its impact on BEV synthesis remained unexplored \cite{bib11}. Similarly, TiledSoilingNet was developed as a granular soiling detection model for embedded systems, but it focused solely on camera views \cite{bib12}. Soiling annotation quality was improved through an ensemble-based semi-supervised learning approach \cite{bib13}, and DirtyGAN, a GAN-based augmentation technique, was introduced, enhancing soiling detection accuracy by 18\% \cite{ bib14}. Despite these advancements, the effect of soiling on multi-sensor BEV systems, particularly in perception tasks such as vehicle segmentation, remains unaddressed.

\subsection{Multi-Sensor Fusion for BEV Perception}
The authors of \cite{bib15} explored multi-sensor fusion with the Simple-BEV framework, which integrates data from cameras, radar, and, optionally, LiDAR to enhance BEV perception. The model processes these inputs through a shared backbone, lifts 2D features into a 3D space, and then fuses this information with radar or LiDAR data to create a robust BEV representation. Radar data notably improves performance, bridging the gap between camera-only and LiDAR-enabled systems. The key focus of our research is to assess the impact of occlusion on the BEV segmentation task. However, prior methods like CVT \cite{bib20}, LSS \cite{bib21}, and CoBEVT \cite{bib22} have enhanced vehicle segmentation accuracy without measuring the impact of environmental factors such as rain, dust, and fog on sensors for BEV perception tasks.

\section{METHODOLOGY}

This section presents our methodology, starting with the creation of occlusion on multi-view cameras on the nuScenes dataset \cite{bib16} to simulate challenging conditions such as moist, rain, and fog. We then apply a multi-sensor fusion technique, using the Simple-BEV \cite{bib15} architecture, which combines radar and LiDAR data with occluded camera inputs to assess the impact of these occlusions on BEV-based perception tasks. While multi-modal approaches have been explored before, our focus is specifically on measuring the degradation caused by soiling and understanding how multi-sensor fusion can mitigate its effects.

\subsection{Dataset}
\begin{figure}[h!]
  \centering
  \includegraphics[width=0.7\columnwidth]{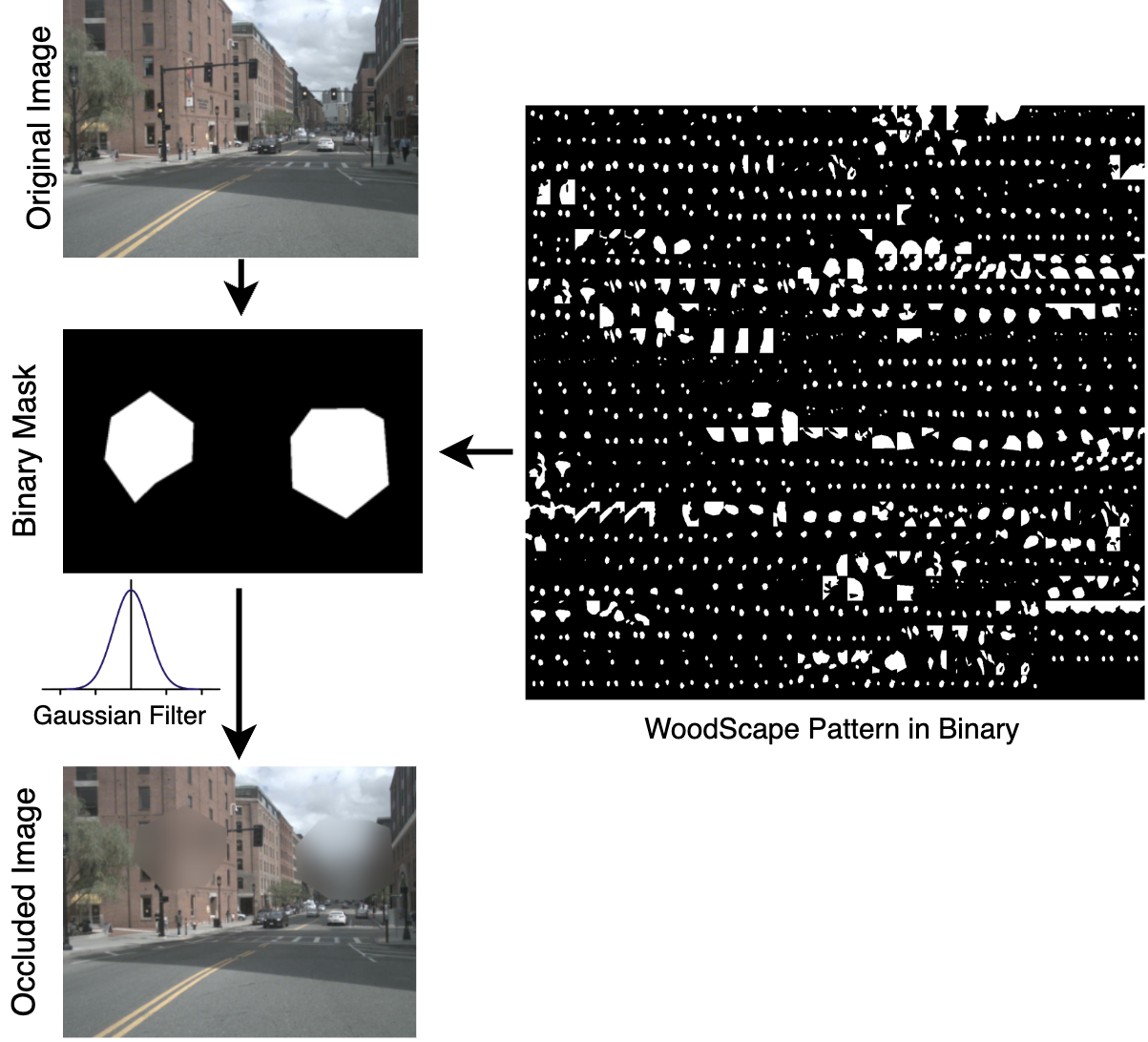}
  \caption{\textbf{The process of generating occlusion patterns from the WoodScape \cite{bib17} Soiled Dataset}: Here 'WoodScape Pattern in Binary' represents various soiling patterns. We then used these patterns and applied a Gaussian filter to "blur it," creating realistic occlusions like "moist" in the final image. }
  \label{Figure:logo1}
\end{figure}

In this work, we use the nuScenes dataset \cite{bib16}, a large-scale dataset for autonomous vehicle research. The dataset features a full sensor suite, including 6 cameras, 5 radars, 1 LiDAR, and GPS \& IMU, providing a comprehensive 360-degree field of view. The dataset contains 1,000 scenes lasting 20 seconds, yielding 1.4 million camera images, 390,000 lidar sweeps, 1.4 million radar sweeps, and 1.4 million annotated 3D bounding boxes across 23 classes. These scenes were captured in diverse environments across Boston and Singapore under various conditions, including nighttime and rainy weather, and include both left-hand and right-hand driving scenarios. The Full nuScenes dataset provides 28,130 samples for training, with approximately 6,019 samples for validation.\\
To enhance the model robustness, we applied artificial occlusions to the nuScenes multi-view camera images. Soiling patterns from the WoodScape Soiled Dataset \cite{bib17} were used to simulate real-world occlusion patterns such as rain, moist, and fog, as depicted in figure \ref{Figure:logo1}. 

 % \vspace*{4mm}
 
A binary mask was generated from the soiling patterns to isolate the occluded areas. A Gaussian filter with a kernel size of (251x251) was then applied to blur only the occluded sections, mimicking the real-world impact of moist or fog. The rest of the image remains unchanged, preserving its original clarity. This augmentation technique simulates autonomous systems' visual challenges in adverse weather and environmental conditions. As seen in figure \ref{Figure:logo1}, "Woodscape patterns in Binary" represents the variety of pattern shapes applied to enhance the occlusion on the nuScenes dataset.

\subsection{ARCHITECTURE OVERVIEW}

\begin{figure*}[t]
    \centering
   % \captionsetup{font={normalsize}, textfont=sf} 
    \includegraphics[width=0.99\textwidth]{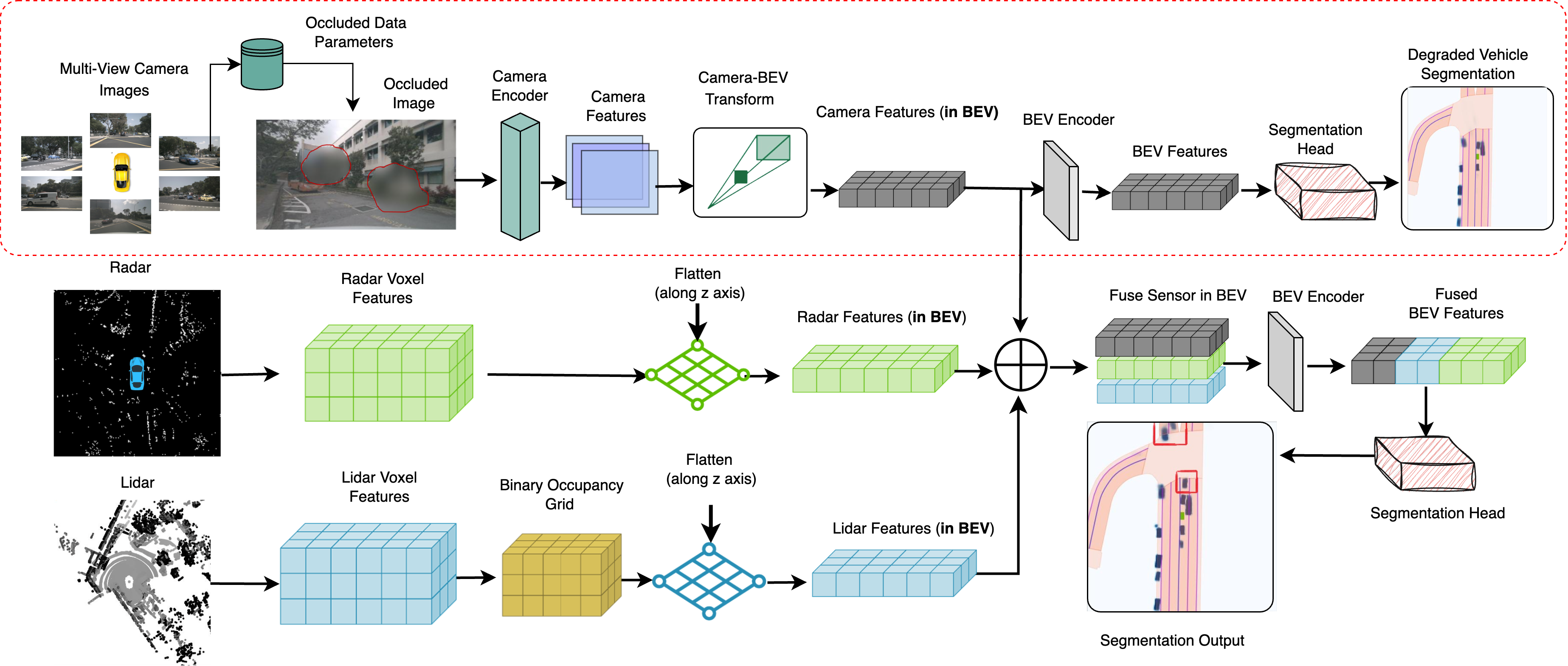}
    \vspace*{4mm}
    \caption{\textbf{Architecture Flow for Simple-BEV:} The diagram illustrates the integration of multi-view camera, radar, and LiDAR sensor data for vehicle segmentation. Occluded data from the cameras is processed and transformed into Bird's-Eye View (BEV) features. These BEV features, along with radar and LiDAR features, are fused to enhance perception and mitigate the effects of occlusion. The model leverages the combined sensor data to improve vehicle segmentation accuracy, as seen in "Segmentation Output," restoring missing information in occluded areas, as highlighted by the red dotted box in the occluded camera pipeline.}
    
    \label{fig:image3}
\end{figure*}

In this architecture, as illustrated in figure \ref{fig:image3}, we used the simple-BEV model \cite{bib15} to evaluate the impact of occlusions on camera lenses, and how these obstructions affect perception tasks, particularly vehicle segmentation in Bird's Eye View (BEV) systems. By employing multi-sensor fusion, we overcome the degradation effects caused by occlusion. First, to establish the baseline of simple-BEV architecture with multi-sensor fusion, we trained the Camera + Radar + Lidar to obtain the weights as they were not provided by simple-BEV.

The architecture begins by feeding multi-view camera images from the nuScenes dataset \cite{bib16}, then we apply the soiling patterns from the WoodScape dataset \cite{bib17}, as detailed in the dataset section. These occluded images are then processed through the Simple-BEV architecture. The first step involves a camera encoder (ResNet-101), which extracts 2D visual features from each camera image, generating feature maps. These features are then processed through convolution layers and upsampled to retain high-resolution spatial information. Next, the model applies bilinear sampling to lift the 2D image features onto the 3D Bird’s Eye View (BEV) plane, converting the camera’s perspective into a top-down view and enabling robust perception for tasks like vehicle segmentation.

Once image features are transformed into BEV space, the features are passed to the BEV encoder (ResNet-18), which encodes the spatial and contextual information into a latent feature map. These features are then passed into the segmentation head, which identifies distinct objects such as vehicles. The final output is a degraded vehicle segmentation in BEV due to occluded input camera images. Hence, the camera-only model’s performance degraded, as depicted in figure \ref{fig:image3} with a red dotted box.

Once we observed the significant degradation with occluded multi-view cameras, we then used multi-sensor fusion radar and LiDAR along with occluded multi-view camera data. The radar and LiDAR provide complementary data, with radar offering information about the velocity of objects and LiDAR delivering precise 3D depth details. These data are processed separately, where radar and LiDAR point clouds are voxelized. For LiDAR data, a binary occupancy grid is applied, where each voxel is assigned a value of 1 if there is at least one LiDAR point in the grid, and 0 otherwise. These voxelized features are then flattened along the z-axis and converted into a 2D BEV format for further processing. The radar and LiDAR features are then merged with occluded camera features in the BEV space to create a more accurate and complete understanding of the surroundings. This fusion allows the system to compensate for the limitations of individual sensors, particularly during camera occlusion.

The fused data is processed through the BEV encoder, which helps mitigate alignment mismatches between the different modalities, ensuring enhanced environmental perception under occluded conditions. Finally, the BEV fused features are passed to the segmentation head, which provides the final segmentation output, representing the accurate segmentation of vehicles.
In general, camera data can identify occlusions, but it may not fully capture the scene behind the occluded areas. Using multi-sensor fusion with the occluded camera ensures a more reliable understanding of the surroundings, as shown in the model's final output with red boxes.

To optimize the model, we employed a Binary Cross Entropy with Logits loss function (BCEWithLogitsLoss) \cite{bib15} for the segmentation head, formulated as:

% \begin{equation}
% \text{BCEWithLogitsLoss} = - \frac{1}{N} \sum_{i=1}^{N} \left[ y_i \cdot \log(\sigma(x_i)) + (1 - y_i) \cdot \log(1 - \sigma(x_i)) \right]
% \end{equation}

\begin{equation}
\begin{aligned}
\text{BCEWithLogitsLoss} = & - \frac{1}{N} \sum_{i=1}^{N} \left[ y_i \cdot \log(\sigma(x_i)) \right. \\
& \left. + (1 - y_i) \cdot \log(1 - \sigma(x_i)) \right]
\end{aligned}
\end{equation}

where \( x_i \) are the predicted logits, \( y_i \) are the ground truth labels, and \( \sigma(x_i) = \frac{1}{1 + e^{-x_i}} \) is the sigmoid function. This function is designed to measure the accuracy of the predicted segmentation occlusion mask against the ground truth labels. This allows the model to learn effectively from the segmentation tasks, including occlusion segmentation in BEV space.

\section{IMPLEMENTATION DETAILS}

In our research, we used the Simple-BEV \cite{bib15} model to train a multi-sensor fusion setup (Camera + Radar + LiDAR) using the PyTorch framework to develop a baseline. We followed the Simple-BEV parameter configuration, utilizing the AdamW optimizer with a learning rate of 3e-4. The model was trained for 80 epochs with an input resolution of 448x800 and a batch size of 12.

\begin{figure*}[t]
    \centering
   % \captionsetup{font={normalsize}, textfont=sf} 
    \includegraphics[width=0.99\textwidth]{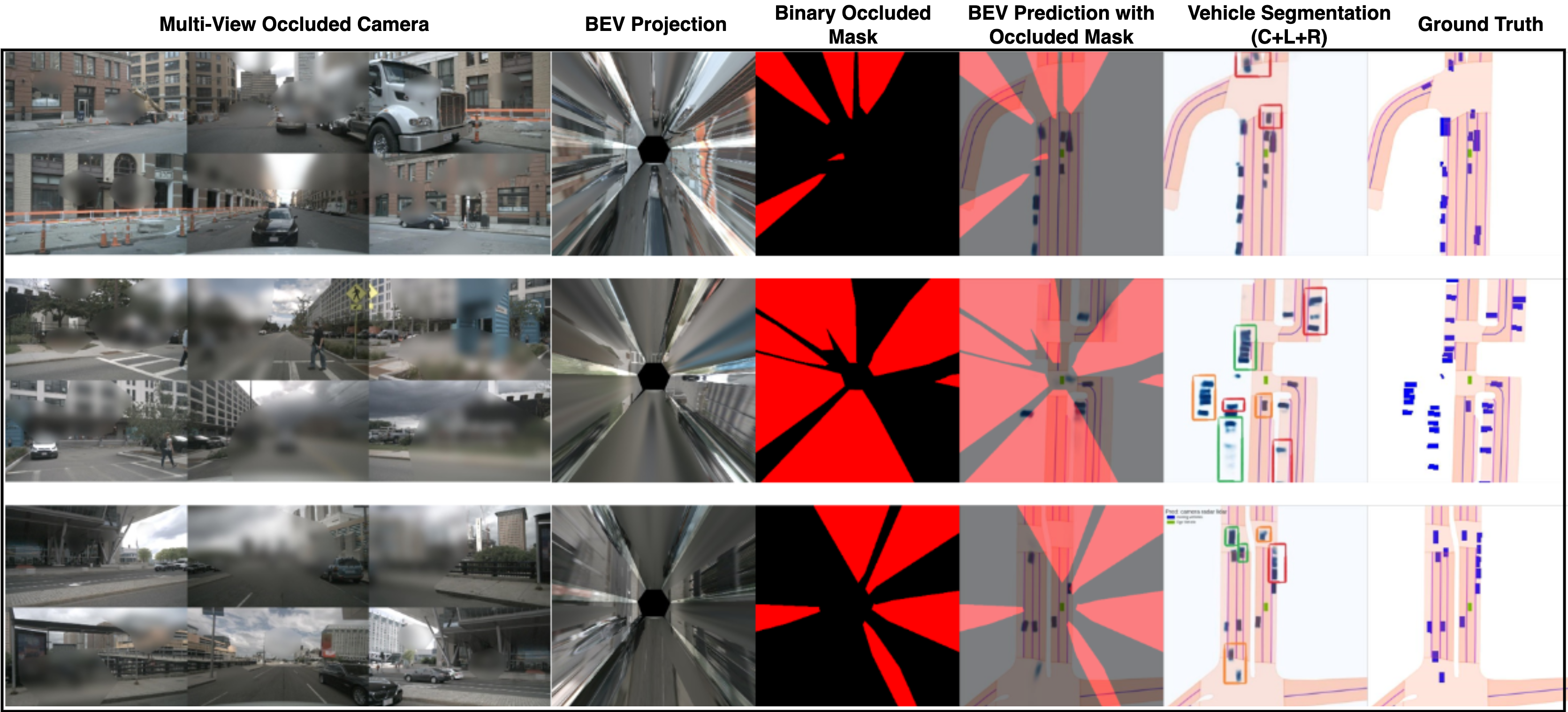}
    \vspace*{4mm}
     \caption{\textbf{Qualitative results:} The sequence from left to right illustrates the multi-view cameras affected by occlusions, followed by the bird's-eye view (BEV) projection of the occluded cameras. Next, we have a binary mask indicating the areas of occlusion in red. Following that is the BEV prediction of the camera with the occluded mask, with missing vehicles due to occlusion. Afterward, we see improved vehicle segmentation using multi-sensor fusion, shown in color boxes, where the missing information is restored. Finally, the ground truth for vehicle segmentation is shown.} 
    \label{fig:image4}
    
\end{figure*}

\section{EXPERIMENTAL RESULTS}

In this section, we discuss the impact of occluded camera images in BEV space. We experiment with three types of occlusion: random box occlusion, overlap region occlusion, and realistic WoodScape pattern occlusion \cite{bib17}. We analyze both quantitative and qualitative results, highlighting how multi-sensor fusion (Radar and LiDAR) improves performance by mitigating the effects of occlusion.

\subsection{Quantitative Analysis}
\begin{figure}[h!]
  \centering
  \includegraphics[width=0.99\columnwidth]{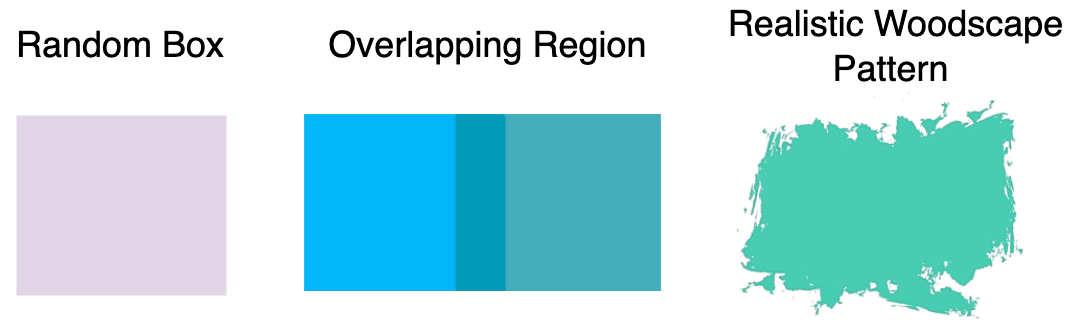}
  \caption{\textbf{Types of occlusions} Left: Random Box occlusion. Middle: Overlapping region occlusion. Right: Realistic Woodscape Pattern.}
  \label{Figure:logo1www}
\end{figure}

In this section, we present the quantitative analysis of our study by evaluating the impact of different types of occlusion on vehicle segmentation as shown in figure \ref{Figure:logo1www}. Also, we discuss the impact of occlusion in table \ref{tab:table1}, using the Intersection over Union (IoU) metric. IoU is a widely recognized metric for evaluating segmentation and detection models, quantifying the overlap between predicted and ground truth bounding boxes, particularly in BEV space.

\begin{table*}[h!]
\centering
\renewcommand{\arraystretch}{1}
% \resizebox{\columnwidth}{!}{
\begin{tabular}{l|llrcrc}
\hline
&&\multicolumn{2}{c}{\textbf{Sensor Modality}}\\
\cline{2-5}
\textbf{Baseline Architecture} \cite{bib15}  & C & C+R  & C+L  & \textbf{C+R+L} \\
\hline
Simple-BEV  & 47.4  & 55.7 & 60.8 & \textbf{64.5}\\
\hline \\
Simple-BEV + \underline{Random}  & 40.6  & 49.6  & 57.3  & 58.3\\
Simple-BEV + \underline{Overlap}  & 45.3  & 54.2  & 56.2  & 62.8\\
Simple-BEV + \underline{Realistic}  & 34.3  & 43.1  & 50.3  & 54.5\\
\hline
Simple-BEV + \underline{Realistic} (Degradation \%) & \textbf{27.6} & \textbf{22.6} & \textbf{17.2}& \textbf{15.5}\\\hline
\end{tabular}
% }
\vspace*{3mm}

\caption{\textmd{\textbf{Table 1: Comparison of Baseline Architecture with Different Types of Occlusion Across Various Sensor Modalities for Vehicle Segmentation}: The values in the table represent the Intersection over Union (IoU) for vehicle segmentation. The Degradation \% compares Simple-BEV with Simple-BEV + Realistic. We observe that as additional sensors are used alongside an occluded camera, the percentage of degradation decreases, indicating that other sensors help mitigate the degradation.}}
\vspace*{3mm}
\label{tab:table1}
\end{table*}

We use the Simple-BEV architecture as the baseline for our experiments, as shown in table \ref{tab:table1}. The reported IoU for Simple-BEV were 47.4 for camera (C) only, 55.7 for camera + radar (C+R), and 60.8 for camera + LiDAR (C+L). However, pre-trained weights for multi-sensor fusion (C+R+L) were not provided, so we trained the setup ourselves and achieved 64.5 IoU.

After establishing the baseline, we evaluate the impact of various types of occlusion on the model's performance, as shown in figure \ref{Figure:logo1www}. First, random square boxes in terms of placement on the multi-view cameras were used to evaluate the impact, followed by occlusion on overlapping regions between camera pairs (e.g., CAM\_FRONT\_LEFT and CAM\_FRONT). Lastly, we evaluated the impact of realistic occlusion patterns from the woodscape soiling dataset \cite{bib17}.

With random box occlusion as shown in figure \ref{Figure:logo1www}, where a square box is randomly placed on each image from the multi-view cameras, performance degraded notably compared to the Simple-BEV baseline. For the camera-only (C) setup, the IoU dropped from 47.4 to 40.6. In the camera + radar (C+R) setup, the IoU fell from 55.7 to 49.6, while in the camera + LiDAR (C+L) setup, it decreased from 60.8 to 57.3. The camera + radar + LiDAR (C+R+L) setup experienced a drop from 64.5 to 58.3.

When occluding overlapping regions as shown in figure \ref{Figure:logo1www}, where occlusion is applied to the overlapping fields of view between camera pairs (for example, CAM\_FRONT\_LEFT and CAM\_FRONT share a common field of view in both images). The performance drop was less severe. The IoU for the camera-only (C) setup dropped to 45.3, while the C+R, C+L, and C+R+L setups saw smaller reductions.

Realistic occlusion caused the most significant degradation. Various patterns and shapes were taken from the Woodscape soiled dataset, converted into binary, and then blurred to apply realistic occlusion. For the camera-only (C) setup, the IoU dropped to 34.3, while for the C+R, C+L, and C+R+L setups, the IoU fell to 43.1, 50.3, and 54.5, respectively.

Random box occlusion obstructs camera views without always targeting vehicle pixels in an image, causing moderate degradation. Overlap occlusion affects only common fields of view between two cameras, resulting in less impact. Realistic occlusion patterns, as shown in figure \ref{Figure:logo1}, mimicking conditions like fog or moisture on the lens, result in the most degradation.

Realistic occlusion caused the most degradation, so we focused on it in our analysis. Using only occluded cameras led to a significant performance loss in terms of vehicle segmentation accuracy. However, using radar and LiDAR sensors reduced this effect. The degradation was notably less in the C+R+L setup, demonstrating how multi-sensor fusion helps overcome occlusion challenges.

\subsection{Qualitative Analysis}
For the qualitative findings, we examined how occlusion affects vehicle segmentation in BEV space and demonstrated how multi-sensor fusion improves segmentation performance by mitigating degradation caused by occlusion in cameras.\\

Figure \ref{fig:image4} illustrates the effect of occlusion in BEV on perception tasks such as vehicle segmentation across multi-view cameras from the NuScenes dataset. The first column shows the multi-view cameras with realistic occlusion patterns. The second and third columns display the bird's-eye view (BEV) projection and the corresponding binary mask of occlusion. The fourth column shows the BEV vehicle segmentation predictions with occluded mask, highlighting where significant degradation in segmentation is observed. As seen, under red occluded areas vehicles were lost due to occlusion. The fifth column demonstrates the improvement achieved through multi-sensor fusion, where information from multiple sensors (occluded camera, radar, and LiDAR) are combined to significantly reduce the negative impact of occlusion, leading to more accurate segmentation results, as highlighted using multi-color boxes. Finally, the last column presents the vehicle segmentation ground truth (GT).\\ 

\section{CONCLUSION}
To conclude, we address a significant gap in the existing literature by focusing on the effects of camera occlusions on BEV-based perception systems in autonomous driving. In this research, we create an occluded version of the nuScenes dataset by applying occlusions to multi-view camera images. We highlight the limitations of camera-only systems by analyzing how these occlusions affect vehicle segmentation accuracy in BEV. Furthermore, our study demonstrates that a multi-sensor fusion approach, which integrates data from radar and LiDAR sensors, successfully mitigates the impact of these occlusions, resulting in improved accuracy and reliability in BEV perception tasks.

\section{FUTURE WORK}
In our future work, we plan to explore the occlusion of cameras under adverse weather conditions, such as rain, night, etc. Additionally, we aim to investigate the impact of other sensor degradation, including occlusion on radar when cameras are clean, and the effects of occlusion on LiDAR when both radar and camera are clean. Furthermore, we intend to assess how these sensor degradations influence tasks that rely on temporal information, such as trajectory forecasting, where temporal clues are crucial for accurate performance. We will also analyze different levels of occlusions, such as opaque and transparent, to better understand their varying impacts on perception accuracy.

\section{ACKNOWLEDGMENTS}
This work was supported with the financial support of the Science Foundation Ireland grant 13/RC/2094\_P2 and co-funded under the European Regional Development Fund through the Southern \& Eastern Regional Operational Programme to Lero - the Science Foundation Ireland Research Centre for Software \href{www.lero.ie}{(www.lero.ie)}

%%%%%%%%%%%%%%%%%%%%%%%%%%%%%%%%%%
% Submitting Your Paper
%%%%%%%%%%%%%%%%%%%%%%%%%%%%%%%%%%

%%%%%%%%%%%%%%%%%%%%%%%%%%%%%%%%%%
% Reference Preparation
%%%%%%%%%%%%%%%%%%%%%%%%%%%%%%%%%%

%\section{Acknowledgments} 
%add the acknowledgement section here


%%%%%%%%%%%%%%%%%%%%%%%%%%%%%%%%%%
% Bibliography
%%%%%%%%%%%%%%%%%%%%%%%%%%%%%%%%%%


\small
\begin{thebibliography}{9}

\bibitem{bib1} M. M. Rana and K. Hossain, "Connected and autonomous vehicles and infrastructures: A literature review," \textit{International Journal of Pavement Research and Technology}, vol. 16, no. 2, pp. 264--284, Springer, 2023.

\bibitem{bib2} J. Zhao, W. Zhao, B. Deng, Z. Wang, F. Zhang, W. Zheng, W. Cao, J. Nan, Y. Lian, and A. F. Burke, "Autonomous driving system: A comprehensive survey," \textit{Expert Systems with Applications}, p. 122836, Elsevier, 2023.

\bibitem{bib3} H. Shao, L. Wang, R. Chen, H. Li, and Y. Liu, "Safety-enhanced autonomous driving using interpretable sensor fusion transformer," in \textit{Conference on Robot Learning}, PMLR, 2023, pp. 726--737.

\bibitem{bib4}
Z. Zou, K. Chen, Z. Shi, Y. Guo, and J. Ye, "Object detection in 20 years: A survey," \textit{Proceedings of the IEEE}, vol. 111, no. 3, pp. 257--276, 2023.

\bibitem{bib5} S. Sharma, A. Das, G. Sistu, M. Halton, and C. Eising, "BEVSeg2TP: Surround View Camera Bird’s-Eye-View Based Joint Vehicle Segmentation and Ego Vehicle Trajectory Prediction," in \textit{Proceedings of the 19th International Joint Conference on Computer Vision, Imaging and Computer Graphics Theory and Applications - Volume 4: VISAPP}, 2024, pp. 25--34.

\bibitem{bib6} Y. Liu, J. Yan, F. Jia, S. Li, A. Gao, T. Wang, and X. Zhang, "Petrv2: A unified framework for 3d perception from multi-camera images," in \textit{Proceedings of the IEEE/CVF International Conference on Computer Vision}, 2023, pp. 3262--3272.

\bibitem{bib7} S. Yao, R. Guan, X. Huang, Z. Li, X. Sha, Y. Yue, E. G. Lim, H. Seo, K. L. Man, X. Zhu, \textit{et al.}, "Radar-camera fusion for object detection and semantic segmentation in autonomous driving: A comprehensive review," \textit{IEEE Transactions on Intelligent Vehicles}, IEEE, 2023.

\bibitem{bib8} Y. Zhang, A. Carballo, H. Yang, and K. Takeda, "Perception and sensing for autonomous vehicles under adverse weather conditions: A survey," \textit{ISPRS Journal of Photogrammetry and Remote Sensing}, vol. 196, pp. 146--177, Elsevier, 2023.

\bibitem{bib9} M. Uřičář, J. Uličný, G. Sistu, H. Rashed, P. Křížek, D. Hurych, A. Vobecký, and S. Yogamani, "Desoiling dataset: Restoring soiled areas on automotive fisheye cameras," in \textit{Proceedings of the IEEE/CVF International Conference on Computer Vision Workshops}, IEEE, 2019, pp. 0--0.


\bibitem{bib10} M. Uřičář, P. Křížek, G. Sistu, and S. Yogamani, "SoilingNet: Soiling detection on automotive surround-view cameras," in \textit{2019 IEEE Intelligent Transportation Systems Conference (ITSC)}, IEEE, 2019, pp. 67--72.

\bibitem{bib11} L. Yahiaoui, M. Uřičář, A. Das, and S. Yogamani, "Let the sunshine in: Sun glare detection on automotive surround-view cameras," in \textit{Electronic Imaging}, vol. 2020, no. 16, Society for Imaging Science and Technology, 2020, pp. 80--1.


\bibitem{bib12} A. Das, P. Křížek, G. Sistu, F. Bürger, S. Madasamy, M. Uřičář, V. R. Kumar, and S. Yogamani, "Tiledsoilingnet: Tile-level soiling detection on automotive surround-view cameras using coverage metric," in \textit{2020 IEEE 23rd International Conference on Intelligent Transportation Systems (ITSC)}, IEEE, 2020, pp. 1--6.


\bibitem{bib13} M. Uřičář, G. Sistu, L. Yahiaoui, and S. Yogamani, "Ensemble-based semi-supervised learning to improve noisy soiling annotations in autonomous driving," in \textit{2021 IEEE International Intelligent Transportation Systems Conference (ITSC)}, IEEE, 2021, pp. 2925--2930.


\bibitem{bib14} M. Uřičář, G. Sistu, H. Rashed, A. Vobecký, V. R. Kumar, P. Křížek, F. Bürger, and S. Yogamani, "Let's get dirty: GAN based data augmentation for camera lens soiling detection in autonomous driving," in \textit{Proceedings of the IEEE/CVF Winter Conference on Applications of Computer Vision}, 2021, pp. 766--775.


\bibitem{bib15} A. W. Harley, Z. Fang, J. Li, R. Ambrus, and K. Fragkiadaki, "Simple-BEV: What really matters for multi-sensor BEV perception," in \textit{2023 IEEE International Conference on Robotics and Automation (ICRA)}, IEEE, 2023, pp. 2759--2765.



\bibitem{bib16} H. Caesar, V. Bankiti, A. H. Lang, S. Vora, V. E. Liong, Q. Xu, A. Krishnan, Y. Pan, G. Baldan, and O. Beijbom, "nuScenes: A multimodal dataset for autonomous driving," in \textit{Proceedings of the IEEE/CVF Conference on Computer Vision and Pattern Recognition}, 2020, pp. 11621--11631.


\bibitem{bib17} S. Yogamani, C. Hughes, J. Horgan, G. Sistu, P. Varley, D. O'Dea, M. Uřičář, S. Milz, M. Simon, K. Amende, \textit{et al.}, "WoodScape: A multi-task, multi-camera fisheye dataset for autonomous driving," in \textit{Proceedings of the IEEE/CVF International Conference on Computer Vision}, 2019, pp. 9308--9318.




\bibitem{bib18} S. Sharma, A. Das, G. Sistu, M. Halton, and C. Eising, "BEVSeg2GTA: Joint Vehicle Segmentation and Graph Neural Networks for Ego Vehicle Trajectory Prediction in Bird’s-Eye-View," \textit{IEEE Access}, 2024, pp. 1--1, doi: 10.1109/ACCESS.2024.3459595.



\bibitem{bib19} S. Hayes, S. Sharma, and C. Eising, "Velocity Driven Vision: Asynchronous Sensor Fusion Birds Eye View Models for Autonomous Vehicles," \textit{arXiv preprint arXiv:2407.16636}, 2024.


\bibitem{bib20}
Zhou, B., \& Kr{\"a}henb{\"u}hl, P. (2022). Cross-view transformers for real-time map-view semantic segmentation. In \textit{Proceedings of the IEEE/CVF Conference on Computer Vision and Pattern Recognition} (pp. 13760--13769).

\bibitem{bib21}
Philion, J., \& Fidler, S. (2020). Lift, splat, shoot: Encoding images from arbitrary camera rigs by implicitly unprojecting to 3D. In \textit{Computer Vision--ECCV 2020: 16th European Conference, Glasgow, UK, August 23--28, 2020, Proceedings, Part XIV} (pp. 194--210). Springer.

\bibitem{bib22}
Xu, R., Tu, Z., Xiang, H., Shao, W., Zhou, B., \& Ma, J. (2022). CoBEVT: Cooperative bird's eye view semantic segmentation with sparse transformers. \textit{arXiv preprint arXiv:2207.02202}.



\end{thebibliography}
\end{document}